\documentclass{article}
\usepackage{spconf,amsmath,graphicx}
\usepackage{enumitem}
\setlist{nosep, leftmargin=14pt}

\usepackage{mwe} 
\usepackage{float}
\usepackage{placeins}
\usepackage{multirow}

\title{Continual pretraining Prostate MRI ViT Encoder for task alignment}
\name{Bilal A. Sidiqi$^{1,2}$, Yipei Wang$^{1,2}$, Shaheer U. Saeed$^{1,2}$, Shiqi Huang$^{1,2}$, Daniel C. Alexander$^{2,3}$, Yipeng Hu$^{1,2}$
}

\address{
    $^{1}$Department of Medical Physics and Biomedical Engineering, \\
University College London, London, UK \\
$^{2}$Hawkes Institute, University College London, London, UK  \\
$^{3}$Department of Computer Science, University College London, London, UK  \\
}
\begin{document}
%
\maketitle
\begin{abstract}

Building foundation models for MRI has emerged as a promising direction for advancing medical image analysis and improving accessibility of task specific models. Existing approaches primarily focus on effective pretraining strategies and evaluating their performance across diverse downstream tasks. While these methods show performance gains, limited attention has been given to understanding task alignment between pretraining and downstream objectives. Unlike language models, where the two stages are naturally aligned, such alignment is less evident in vision based applications. Motivated by this gap, we explore continual pretraining. Specifically, continually pretraining a pretrained ViT encoder using Low-Rank Adaptation (LoRA) on a prostate segmentation task to enhance task alignment and downstream performance. We further perform layer-wise representational analysis using Centered Kernel Alignment (CKA) to assess representational shifts and their correlation with task outcomes. Our experiments show improved performance in 3 of 6 downstream tasks and comparable results in the others. Our Feature analysis reveals that, for segmentation tasks, continually pretrained encoder features exhibit greater similarity to those of the finetuned encoder than the original pretrained encoder, correlating with observed performance gains.


\end{abstract}
\section{Introduction}
\label{sec:intro}

Prostate MRI has an important role in reducing number of biopsies and damage to surrounding tissues during clinical diagnosis \cite{9461197}. With the help of AI models, automation of information extraction, such as lesion segmentation, can further enhance clinical decision making \cite{turkbey2022deep}. Traditionally, independent models have been developed for different tasks, which is resource intensive, limits practical deployment, and hinders generalisation to out-of-distribution data \cite{wang2024sam}. Motivated by the success of large language models (LLMs), recent efforts to develop large scale medical vision models have shown considerable promise but continue to face several limitations. Notably, features learned through both general-purpose \cite{Wu2023MedicalSA} and domain-specific \cite{Burshtein2025CXRDINOPT} pretraining have not consistently translated to strong performance in medical downstream tasks, often requiring extensive task-specific fine-tuning. Although various pretraining strategies such as MAE, DINO, and CLIP have been explored \cite{campanella2023computational,Zhao2023CLIPIM}, each offering task-dependent advantages, they still seem to lack balanced representations for capturing the shared semantics across downstream tasks \cite{ilse2025data}. 


In this work, we propose a continual pretraining approach aimed at improving the learning of shared downstream task representations. Continual pretraining generally involves further pretraining an already pretrained encoder to better align its learned features with those relevant to downstream tasks. A recent study \cite{ilse2025data} explored a similar idea by continually pretraining self-supervised DINOv2 and supervised CLIP models on domain-specific data at scales of 30k, 50k, 100k, 1M, and 1.3M samples, reporting consistent performance improvements with larger in-domain datasets—where, in some cases, as few as 30k samples outperformed the original pretrained weights. While our dataset is considerably smaller, we nevertheless observe improved performance in several in-domain and out-of-domain downstream tasks.

Specifically, we employ a pretrained Vision Transformer (ViT) encoder \cite{dosovitskiy2020image} trained using 5,524 3D prostate MRI scans on a self-supervised Masked Autoencoder (MAE) objective to capture multi-level general features from MRI data.  We continually pretrain this encoder using Low-Rank Adaptation (LoRA) \cite{Hu2021LoRALA} on a prostate segmentation task comprising 396 training and 113 test samples. The prostate segmentation task was chosen due to its close relevance to several downstream tasks, including prostate lesion and anatomical segmentation, PI-RADS score classification, and prostate volume regression. Across six downstream task examples, we observe performance improvements in three tasks following continual pretraining. Two of which involve segmentation, highlighting the significance of task similarity for achieving better alignment during continual pretraining.

To further analyse representational changes, we utilise the Centered Kernel Alignment (CKA) similarity metric \cite{Kornblith2019SimilarityON} to quantify pairwise feature representation similarities across three encoders: the pretrained baseline, the fine-tuned model (initialised from the pretrained encoder), and our continual pretrained encoder. We analyse the difference between CKA(pretrained, fine-tuned) and CKA(continual-pretrained, fine-tuned) to assess whether continual pretraining shifts the feature geometry closer to that of the fine-tuned model, indicating improved task alignment. Furthermore, we examine whether the change in representational similarity correlates with the downstream performance difference between fine-tuned models initialised from pretrained and continually pretrained encoders. Our results show that this correlation holds when a measurable representational similarity shift is observed, suggesting that geometric feature alignment is indeed important in improved downstream performance.

\section{Method}
\label{sec:method}

\subsection{Applying LoRA to a Pretrained Encoder}
We propose a continual pretraining framework in which a 3D Vision Transformer (ViT) encoder, pretrained on a self-supervised Masked Autoencoder (MAE) objective, is further aligned to downstream tasks using Low-Rank Adaptation (LoRA) through supervised prostate segmentation. Specifically, LoRA is integrated into the attention mechanism of each transformer block within the encoder. This modifies the original query, key, value, and output projection matrices $(W_Q, W_K, W_V, W_O)$ by introducing trainable low-rank update matrices, formulated as:
\begin{align}
W = W_0 + \alpha BA,
\end{align}
where $W_{0\in\{Q,K,V,O\}}$ represents the original weight matrix, $A$ and $B$ are low-rank matrices capturing essential modes of variation within the weight space, and $\alpha$ is a scaling factor controlling the adaptation magnitude. The adapted transformer block can be expressed as:
{\footnotesize
\begin{align}
&\hat{Z}^{(\ell)} = \mathrm{LayerNormalisation}\!\big(Z^{(\ell-1)}\big), \\
&Q = \hat{Z}^{(\ell)} W_Q, \quad 
  K = \hat{Z}^{(\ell)} W_K, \quad 
  V = \hat{Z}^{(\ell)} W_V, \\
&\mathrm{Attention}(Q,K,V) =
   \mathrm{softmax}\!\left(\frac{QK^\top}{\sqrt{d_h}}\right)V, \\
&Z^{(\ell,1)} = 
   Z^{(\ell-1)} + \mathrm{Attention}(Q,K,V) W_O, \\
&Z^{(\ell)} = Z^{(\ell,1)} + \mathrm{MLP}\Big(
\mathrm{LayerNormalisation}\big(Z^{(\ell,1)}\big)\Big),
\end{align}
}

where $\hat{Z}^{(\ell)}$ denotes the normalized layer output, $d_h$ is the dimensionality of the query/key vectors, $W_{0\in\{Q,K,V,O\}}$ are the LoRA-adapted weight matrices and MLP is Multilayer Perceptron.

\subsection{Continual Pretraining}
During continual pretraining, the pretrained encoder weights $W$ are frozen, and an adapter module is used to connect a Unified Perceptual Parsing (UperNet) decoder for segmentation. The adapter receives hidden states $\{H_i\}_{i=1}^4$ from selected layers of the ViT encoder and reshapes each token sequence into a 3D feature map $X_i = \mathrm{Reshape}(H_i) \in {R}^{B \times C \times D \times H \times W}$. The UperNet decoder integrates these multiscale feature maps by downsampling the first two and upsampling the last to align them to the spatial scale of the third map. The fused representation is then passed through a softmax classifier to produce voxel-level segmentation. LoRA parameters are trained to adapt the encoder from generating general features learned during pretraining to producing task-aligned representations more suitable for related downstream tasks.

\subsection{Downstream Tasks}
Following continual pretraining, the encoder’s original weights are unfrozen and evaluated across a range of downstream tasks using both in-domain (identical to continual pretraining data) and out-of-domain datasets, as summarised in Table~\ref{tab:datasets}. To isolate encoder effects, we employ simple task specific heads. A lightweight two layer linear head is attached for prostate zone classification and prostate volume regression. For lesion and prostate zone segmentation, we employ the same UperNet decoder architecture. Finally, a super-resolution module from \cite{Dong2014ImageSU} is attached to the encoder to enhance low-resolution $T_2$ inputs by a factor of four.

\subsection{Geometric Feature Analysis Metric}
To analyse layer-wise representational similarity across the pretrained, continual-pretrained, and fine-tuned encoders, we employ the Centered Kernel Alignment (CKA) similarity metric. CKA quantifies the degree to which two sets of feature representations encode similar geometric structures in their respective feature spaces. We compute pairwise CKA values between encoder layers (e.g. $l_1$ and $l_2$) for tasks exhibiting a performance difference of approximately $0.2\%$ or greater. For given layer outputs $Z_1^{(l_1)}$ and $Z_2^{(l_2)}$ from encoders $Z_1$ and $Z_2$, the linear CKA similarity is defined as:
\begin{equation}
\mathrm{CKA}\big(Z_1^{(l_1)}, Z_2^{(l_2)}\big) =
\frac{\| {Z_1^{(l_1)}}^\top Z_2^{(l_2)} \|_F^2}
     {\| {Z_1^{(l_1)}}^\top Z_1^{(l_1)} \|_F \;
      \| {Z_2^{(l_2)}}^\top Z_2^{(l_2)} \|_F},
\end{equation}
where $\|\cdot\|_F$ denotes the Frobenius norm. In our experiments, CKA similarities are computed for layers 3, 5, 7, and 11 across the three encoders to evaluate how fine-tuning and continual pretraining influence internal feature representations relative to the pretrained baseline.

\begin{figure}[h]
    \centering
    \includegraphics[width=1\linewidth]{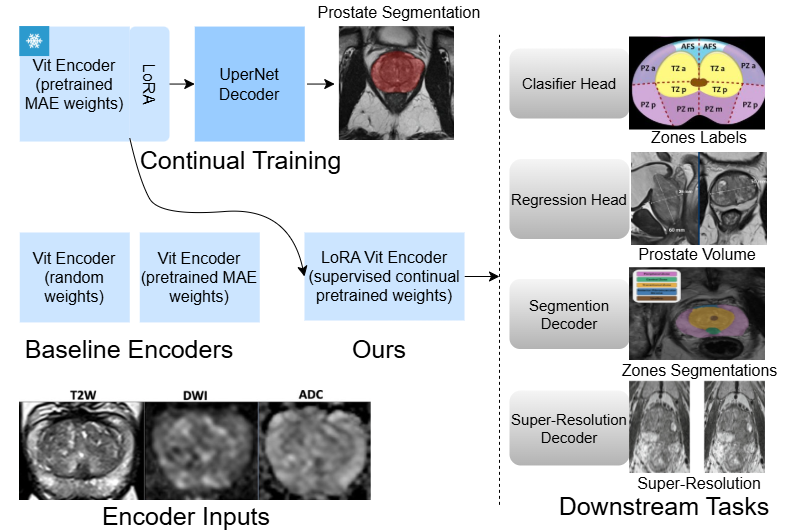}
    \caption{Overview of the proposed continual pretraining framework, integrating LoRA within a pretrained ViT encoder to continually pretrain the encoder for better task alignment and downstream performance.  }
    \label{fig:framework}
\end{figure}



\section{Experiments}
\label{sec:pagestyle}

\subsection{Dataset and Experimental Setup}

Four datasets were used in this study: PROMIS for continual pretraining, Anatomy for downstream anatomy segmentation, RISK for PIRADS score classification, and UCL Set for Prostate volume regression \cite{marsden2022reimagine,LI2023102935,AHMED2017815}. Their respective data splits, input modalities, and target outputs are summarised in Table~\ref{tab:datasets}. All models were trained with an input size of $(64, 256, 256)$, except for the super-resolution task where the depth dimension ($64$) was reduced by a factor of four. The inputs underwent standard preprocessing and augmentation procedures, including normalisation, centre cropping, random rotation, flipping, zooming, and intensity variation.  

All experiments were conducted on an NVIDIA A40 GPU for 100 epochs. The Adam optimiser was used across all tasks, with Dice loss applied to segmentation, cross-entropy loss to classification, mean squared error (MSE) to regression, and mean absolute error (MAE) to super-resolution. Batch sizes were set to $4$ for segmentation and super-resolution tasks, and $16$ for classification and regression tasks.

\begin{table}[h]
  \centering
  \renewcommand{\arraystretch}{1.5}
  \setlength{\tabcolsep}{2pt}
  \small
  \begin{tabular}{|c|c|c|c|}
    \hline
    \textbf{Dataset} & \textbf{Data Split} & \textbf{MRI Input} & \textbf{Output} \\ \hline
    PROMIS  & (396,57,113) & T2W, DWI, ADC & Lesion Mask \\ \hline
    Anatomy & (412,60,117) & T2W & Prostate Zones Mask \\ \hline
    RISK    & (718,103,207) & T2W, DWI, ADC & PIRADS Score (2–5) \\ \hline
    UCL Set & (1386,178,387) & T2W & Prostate Volume \\ \hline
  \end{tabular}
  \caption{Overview of datasets, data splits, and corresponding input–output configurations.}
  \label{tab:datasets}
\end{table}

\subsection{Results}
Model performance was compared across three configurations sharing identical architectures but differing in weight initialisation. The first, \textit{ViT Random Weights}, was trained from scratch. The second, \textit{ProFound ViT OMW}, initialised with pretrained encoder weights, served as the baseline. The third, \textit{ProFound ViT Ours}, utilised our continually pretrained encoder weights.


\noindent\textbf{Performance on Classification Tasks:} Two classification tasks were evaluated. The first employed the RISK dataset for four-class PIRADS score classification (2–5), where higher scores indicate increased malignancy risk. The second used the PROMIS dataset for 20-zone prostate classification. As shown in Table~\ref{tab:cls}, for PIRADS classification, the baseline model achieved the highest overall AUC, outperforming our model by approximately $0.4\%$. However, for AUC thresholds $\geq3$ and $\geq4$, our model surpassed the baseline by over $5\%$ in AUC$\geq3$, suggesting improved discrimination of higher-risk PIRADS scores. For zone classification, the baseline outperformed our model by $2.3\%$ on average.

Feature representation analysis (Section~3.2.5) revealed that encoder representations from the continually pretrained and finetuned models were highly similar to the pretrained encoder. This suggests that for classification tasks, globally learned representations during pretraining remain largely transferable, and performance differences are likely influenced by subtle variations in discriminative feature directions~\cite{article}.


\noindent\textbf{Performance on Regression Task:} As presented in Table~\ref{tab:reg}, regression results showed a marginal $0.2\%$ decrease compared to the baseline, indicating limited benefit of continual pretraining for this specific alignment task.


\noindent\textbf{Performance on Segmentation Tasks:} We evaluated two segmentation tasks: lesion segmentation using the PROMIS dataset and prostate zone segmentation using the Anatomy dataset (eight zones in total). Table~\ref{tab:seg} shows that our model outperformed the baseline by approximately $0.5\%$ in average Dice score for lesion segmentation and $0.1\%$ for zone segmentation. These improvements suggest that continual pretraining on a similar segmentation task enhances alignment for downstream segmentation objectives. This observation is further examined in Section~3.1.1, where representational similarity is analysed.


\noindent\textbf{Performance on Super-Resolution Task:} Table~\ref{tab:sr} presents results for the super-resolution task, where the baseline model achieved a slightly higher SSIM ($0.1\%$ improvement). Similar to the regression case, continual pretraining did not yield improved task alignment for this setting.

\begin{table}[h]
  \centering
  \small
  \setlength{\tabcolsep}{4pt} 
  \renewcommand{\arraystretch}{1.1} 
  \begin{tabular}{|l|c|c|c|c|}
    \hline
    \multicolumn{5}{|c|}{\textbf{PIRADS Classification (RISK Data)}} \\ \hline
    & AUC & AUC $\geq$3 & AUC $\geq$4 & Avg Std \\ \hline
    ViT Random Weights & 58.81 & 69.98 & 62.82 & 2.51 \\ 
    ProFound ViT OMW & \textbf{61.36} & 74.76 & \textbf{64.03} & 2.56 \\ 
    ProFound ViT Ours & 60.94 & \textbf{79.92} & 64.00 & 2.63 \\ \hline
    \multicolumn{5}{|c|}{\textbf{Zone Classification (PROMIS Data)}} \\ \hline
    ViT Random Weights & 64.41 & – & – & 1.97 \\ 
    ProFound ViT OMW & \textbf{72.64} & – & – & 1.75 \\ 
    ProFound ViT Ours & 70.36 & – & – & 1.86 \\ \hline
  \end{tabular}
  \caption{Results for classification tasks.}
  \label{tab:cls}
\end{table}

\begin{table}[h]
  \centering
  \begin{tabular}{|c|c|c|}
    \hline
    \multicolumn{3}{|c|}{\textbf{Prostate Volume Regression (UCL Data)}} \\ \hline
    & MSE & Std\\ \hline
    ViT Random Weights   & 0.038   & 0.028   \\ \hline
    ProFound ViT OMW   & \textbf{0.013}   & 0.009   \\ \hline
    ProFound ViT Ours   & 0.015  & 0.012 \\ \hline
  \end{tabular}
  \caption{Results for regression task.}
  \label{tab:reg}
\end{table}

\begin{table}[h]
\centering
\scriptsize 
\renewcommand{\arraystretch}{1.15} 
\setlength{\tabcolsep}{4pt} 
\begin{tabular}{|c|c|p{1.8cm}|c|}
  \hline
  \multicolumn{4}{|c|}{\textbf{Lesion Segmentation (PROMIS Data)}} \\ \hline
  & \textbf{Avg Dice} & \textbf{Class Dice} & \textbf{Std} \\ \hline
  ViT Random Weights & 0.216 & – & 0.186 \\ \hline
  ProFound ViT OMW & 0.248 & – & 0.203 \\ \hline
  ProFound ViT Ours & \textbf{0.253} & – & 0.206 \\ \hline
  \multicolumn{4}{|c|}{\textbf{Zone Segmentation (Anatomy Data)}} \\ \hline
  & \textbf{Avg Dice} & \textbf{Class Dice} & \textbf{Std} \\ \hline
  ViT Random Weights & 0.830 & 0.910, \textbf{0.913}, \textbf{0.868}, 0.771, 0.875, 0.907, 0.750, 0.646 & 0.093 \\ \hline
  ProFound ViT OMW & 0.835 & 0.916, 0.908, 0.866, \textbf{0.782}, 0.880, 0.912, 0.765, 0.653 & 0.095 \\ \hline
  ProFound ViT Ours & \textbf{0.836} & \textbf{0.917}, 0.909, 0.866, 0.780, \textbf{0.881}, \textbf{0.912}, \textbf{0.765}, \textbf{0.657} & 0.095 \\ \hline
\end{tabular}
\caption{Results for segmentation tasks.}
\label{tab:seg}
\end{table}

\begin{table}[h]
  \centering
  \begin{tabular}{|c|c|c|}
    \hline
    \multicolumn{3}{|c|}{\textbf{T2 MRI Super-Resolution (PROMIS Data)}} \\ \hline
    & SSIM & Std \\ \hline
    ViT Random Weights   & 0.900   & 0.009 \\ \hline
    ProFound ViT OMW   & \textbf{0.931} & 0.005  \\ \hline
    ProFound ViT Ours   & 0.930 & 0.005 \\ \hline
  \end{tabular}
  \caption{Results for super-resolution task.}
  \label{tab:sr}
\end{table}


\noindent\textbf{Feature Representation Analysis:} Table~\ref{tab:cka_summary_compact} presents the layer wise and average CKA similarity scores between encoder pairs across four selected layers (3, 5, 7, and 11). For the pretrained–continual pair, the encoders exhibit high similarity across all layers, with a gradual decrease in deeper layers, suggesting that LoRA primarily refines high-level semantic representations while inducing task-specific adaptations in later layers.

In the lesion segmentation task, finetuned encoder representations (using pretrained weights) show decreasing similarity to both pretrained and continual pretrained encoders with layer depth. Notably, at the deepest layer, similarity with the continual pretrained encoder increases by approximately $0.4\%$, aligning with the observed improvement in segmentation performance (Table~\ref{tab:seg}). This supports the hypothesis that continual pretraining facilitates better task-aligned representation learning.

For the PIRADS classification task, the finetuned encoder remains nearly identical to both pretrained and continual pretrained encoders, implying minimal representational shifts during finetuning. This suggests that classification primarily benefits from globally stable representations rather than layer specific adaptations.

\begin{table}[h]
\centering
\renewcommand{\arraystretch}{1.1}
\setlength{\tabcolsep}{3pt}
\begin{tabular}{|c|cccc|c|}
\hline
\textbf{Encoder Pair ($Z_1-Z_2$)} & L3 & L5 & L7 & L11 & \textbf{Avg.} \\ \hline
Pretrained--Continual & 0.997 & 0.996 & 0.997 & 0.995 & 0.996 \\ \hline
\multicolumn{6}{|c|}{\textbf{CKA Similarity for Lesion Segmentation}} \\ \hline
Pretrained--Finetuned & \textbf{0.918} & \textbf{0.915} & \textbf{0.893} & 0.340 & \textbf{0.767} \\
Continual--Finetuned  & 0.915 & 0.910 & 0.890 & \textbf{0.344} & 0.765 \\ \hline
\multicolumn{6}{|c|}{\textbf{CKA Similarity for PIRADS Classification}} \\ \hline
Pretrained--Finetuned & \textbf{1.000} & \textbf{1.000} & \textbf{1.000} & \textbf{0.993} & \textbf{0.998} \\ 
Continual--Finetuned  & 0.997 & 0.997 & 0.997 & 0.989 & 0.995 \\ \hline
\end{tabular}
\caption{Layer-wise and average CKA similarity between encoder pairs. “Pretrained” denotes the encoder with pretrained weights; “Continual” indicates the pretrained encoder further trained with LoRA on the prostate segmentation task; “Finetuned” refers to the encoder finetuned on downstream tasks (lesion segmentation or PIRADS classification).}
\label{tab:cka_summary_compact}
\end{table}


\section{Conclusion}
In this work, we investigated a novel direction of continual pretraining as a means of advancing foundation model development for prostate MRI analysis.
The objective was to enhance task alignment and improve downstream task performance. Our experimental results demonstrate improved and comparable performances in all 6 downstream tasks, two of which involve segmentation, suggesting that task similarity during continual pretraining contributes to improved alignment.
Our Feature analysis results show that, for segmentation tasks, the output feature representations of the continual pretrained encoder exhibit greater similarity to those of the finetuned (initialised from the pretrained) encoder compared to the pretrained encoder itself. This increased representational closeness correlates with the observed performance improvements when finetuning from continual pretrained weights. Conversely, for classification tasks, the representational geometry exhibited minimal change during finetuning, likely due to a stronger focus on fine grained discriminative details, therefore limiting the ability to infer a clear correlation with performance.


Future work may explore integrating a Mixture-of-Experts (MoE) approach in the later layers of the encoder to facilitate multi-task continual pretraining and achieve more generalisable task alignment. Moreover, employing alternative representational similarity metrics could provide deeper insights into the impact of continual pretraining on feature representations.

\section{COMPLIANCE WITH ETHICAL STANDARDS}
This study was performed in line with the principles of the Declaration of Helsinki. Ethical approval was granted by the Data Trust Committee, University College London Hospitals (24-04-2024/IRAS ID 299136).












\bibliographystyle{IEEEtran}
\bibliography{ref}

@ARTICLE{9461197,
  author={Khan, Zia and Yahya, Norashikin and Alsaih, Khaled and Al-Hiyali, Mohammed Isam and Meriaudeau, Fabrice},
  journal={IEEE Access}, 
  title={Recent Automatic Segmentation Algorithms of MRI Prostate Regions: A Review}, 
  year={2021},
  volume={9},
  number={},
  pages={97878-97905},
  doi={10.1109/ACCESS.2021.3090825}}

@article{turkbey2022deep,
  title={Deep learning-based artificial intelligence applications in prostate MRI: brief summary},
  author={Turkbey, Baris and Haider, Masoom A},
  journal={The British Journal of Radiology},
  volume={95},
  number={1131},
  pages={20210563},
  year={2022},
  publisher={The British Institute of Radiology.}
}

@inproceedings{wang2024sam,
  title={Sam-med3d: towards general-purpose segmentation models for volumetric medical images},
  author={Wang, Haoyu and Guo, Sizheng and Ye, Jin and Deng, Zhongying and Cheng, Junlong and Li, Tianbin and Chen, Jianpin and Su, Yanzhou and Huang, Ziyan and Shen, Yiqing and others},
  booktitle={European Conference on Computer Vision},
  pages={51--67},
  year={2024},
  organization={Springer}
}

@article{Wu2023MedicalSA,
  title={Medical SAM Adapter: Adapting Segment Anything Model for Medical Image Segmentation},
  author={Junde Wu and Rao Fu and Huihui Fang and Yuanpei Liu and Zhao-Yang Wang and Yanwu Xu and Yueming Jin and Tal Arbel},
  journal={Medical image analysis},
  year={2023},
  volume={102},
  pages={
          103547
        },
  url={https://api.semanticscholar.org/CorpusID:258309597}
}

@inproceedings{Burshtein2025CXRDINOPT,
  title={CXR-DINO: paving the way for a medical vision foundation model through self-supervised learning in chest x-ray analysis},
  author={Evyatar Burshtein and Noa Cahan and Lev Ayzenberg and Hayit Greenspan},
  booktitle={Medical Imaging},
  year={2025},
  url={https://api.semanticscholar.org/CorpusID:277756367}
}

@article{campanella2023computational,
  title={Computational Pathology at Health System Scale--Self-Supervised Foundation Models from Three Billion Images},
  author={Campanella, Gabriele and Kwan, Ricky and Fluder, Eugene and Zeng, Jennifer and Stock, Aryeh and Veremis, Brandon and Polydorides, Alexandros D and Hedvat, Cyrus and Schoenfeld, Adam and Vanderbilt, Chad and others},
  journal={arXiv preprint arXiv:2310.07033},
  year={2023}
}

@article{Zhao2023CLIPIM,
  title={CLIP in Medical Imaging: A Comprehensive Survey},
  author={Zihao Zhao and Yuxiao Liu and Han Wu and Yonghao Li and Sheng Wang and Lin Teng and Disheng Liu and Xiang Li and Zhiming Cui and Qian Wang and Dinggang Shen},
  journal={ArXiv},
  year={2023},
  volume={abs/2312.07353},
  url={https://api.semanticscholar.org/CorpusID:266174283}
}

@article{ilse2025data,
  title={Data Scaling Laws for Radiology Foundation Models},
  author={Ilse, Maximilian and Sharma, Harshita and Schwaighofer, Anton and Bond-Taylor, Sam and P{\'e}rez-Garc{\'\i}a, Fernando and Melnichenko, Olesya and Sykes, Anne-Marie G and Horst, Kelly K and Khandelwal, Ashish and Reynolds, Maxwell and others},
  journal={arXiv preprint arXiv:2509.12818},
  year={2025}
}

@article{dosovitskiy2020image,
  title={An image is worth 16x16 words: Transformers for image recognition at scale},
  author={Dosovitskiy, Alexey},
  journal={arXiv preprint arXiv:2010.11929},
  year={2020}
}

@article{Hu2021LoRALA,
  title={LoRA: Low-Rank Adaptation of Large Language Models},
  author={J. Edward Hu and Yelong Shen and Phillip Wallis and Zeyuan Allen-Zhu and Yuanzhi Li and Shean Wang and Weizhu Chen},
  journal={ArXiv},
  year={2021},
  volume={abs/2106.09685},
  url={https://api.semanticscholar.org/CorpusID:235458009}
}

@article{Kornblith2019SimilarityON,
  title={Similarity of Neural Network Representations Revisited},
  author={Simon Kornblith and Mohammad Norouzi and Honglak Lee and Geoffrey E. Hinton},
  journal={ArXiv},
  year={2019},
  volume={abs/1905.00414},
  url={https://api.semanticscholar.org/CorpusID:141460329}
}

@article{Dong2014ImageSU,
  title={Image Super-Resolution Using Deep Convolutional Networks},
  author={Chao Dong and Chen Change Loy and Kaiming He and Xiaoou Tang},
  journal={IEEE Transactions on Pattern Analysis and Machine Intelligence},
  year={2014},
  volume={38},
  pages={295-307},
  url={https://api.semanticscholar.org/CorpusID:6593498}
}

@article{article,
author = {Ruiping, Yang and Kun, Liu and Shaohua, Xu and Jian, Yin and Zhen, Zhang},
year = {2024},
month = {02},
pages = {},
title = {ViT-UperNet: a hybrid vision transformer with unified-perceptual-parsing network for medical image segmentation},
volume = {10},
journal = {Complex \& Intelligent Systems},
doi = {10.1007/s40747-024-01359-6}
}

@article{marsden2022reimagine,
  title={The ReIMAGINE prostate cancer risk study protocol: A prospective cohort study in men with a suspicion of prostate cancer who are referred onto an MRI-based diagnostic pathway with donation of tissue, blood and urine for biomarker analyses.},
  author={Marsden, Teresa and McCartan, Neil and Brown, Louise and Rodriguez-Justo, Manuel and Syer, Tom and Brembilla, Giorgio and Van Hemelrijck, Mieke and Coolen, Ton and Attard, Gerhardt and Punwani, Shonit and others},
  journal={Plos one},
  volume={17},
  number={2},
  pages={e0259672},
  year={2022},
  publisher={Public Library of Science San Francisco, CA USA}
}

@article{LI2023102935,
title = {Prototypical few-shot segmentation for cross-institution male pelvic structures with spatial registration},
journal = {Medical Image Analysis},
volume = {90},
pages = {102935},
year = {2023},
issn = {1361-8415},
doi = {https://doi.org/10.1016/j.media.2023.102935},
author = {Yiwen Li and Yunguan Fu and Iani J.M.B. Gayo and Qianye Yang and Zhe Min and Shaheer U. Saeed and Wen Yan and Yipei Wang and J. Alison Noble and Mark Emberton and Matthew J. Clarkson and Henkjan Huisman and Dean C. Barratt and Victor A. Prisacariu and Yipeng Hu}
}

@article{AHMED2017815,
title = {Diagnostic accuracy of multi-parametric MRI and TRUS biopsy in prostate cancer (PROMIS): a paired validating confirmatory study},
journal = {The Lancet},
volume = {389},
number = {10071},
pages = {815-822},
year = {2017},
issn = {0140-6736},
doi = {https://doi.org/10.1016/S0140-6736(16)32401-1},
author = {Hashim U Ahmed and Ahmed {El-Shater Bosaily} and Louise C Brown and Rhian Gabe and Richard Kaplan and Mahesh K Parmar and Yolanda Collaco-Moraes and Katie Ward and Richard G Hindley and Alex Freeman and Alex P Kirkham and Robert Oldroyd and Chris Parker and Mark Emberton}
}

\end{document}